\documentclass[runningheads]{llncs}
\usepackage[T1]{fontenc}
\usepackage{graphicx}
\usepackage{cite}
\usepackage{amsmath,amssymb,amsfonts}
\usepackage{algorithmic}
\usepackage{textcomp}
\usepackage[table]{xcolor}
\usepackage{tabularx}
\usepackage[colorlinks,allcolors=blue]{hyperref}
\usepackage{ragged2e}
\usepackage{comment}
\usepackage{multirow}
\usepackage{xspace}
\usepackage{booktabs}
\usepackage{mathrsfs}
\usepackage{url}
\usepackage{subcaption}
\usepackage{listings}
\usepackage{caption}
\usepackage{float}      
\usepackage{tikz}
\usepackage[most]{tcolorbox}
\usepackage{enumitem}
\usepackage{wrapfig}
\usepackage{placeins}
\usetikzlibrary{trees}
\usetikzlibrary{shapes.callouts}
\usetikzlibrary{positioning} 

\begin{document}
%
\title{CRASH: Cognitive Reasoning Agent for Safety Hazards in Autonomous Driving}
\titlerunning{CRASH: Cognitive Reasoning Agent for Safety Hazards}
%
%
\author{Erick Silva\inst{1}\orcidID{0009-0003-7352-039X} \and Rehana Yasmin\inst{1}\orcidID{0009-0006-9039-0648} \and
Ali Shoker\inst{1}\orcidID{0000-0002-4898-9394}}
\authorrunning{S. Erick et al.}
%
\institute{King Abdullah University of Science and Technology, Thuwal, Saudi Arabia\\
\email{\{firstname.lastname\}@kaust.edu.sa}}
\maketitle              
\begin{center}
  \small\textit{Preprint --- Currently Under Review}
\end{center}
\begin{abstract}

As AVs grow in complexity and diversity, identifying the root causes of operational failures has become increasingly complex. The heterogeneity of system architectures across manufacturers, ranging from end-to-end to modular designs, together with variations in algorithms and integration strategies, limits the standardization of incident investigations and hinders systematic safety analysis. This work examines real-world AV incidents reported in the NHTSA database. We curate a dataset of 2,168 cases reported between 2021 and 2025, representing more than 80 million miles driven. To process this data, we introduce CRASH, Cognitive Reasoning Agent for Safety Hazards, an LLM-based agent that automates reasoning over crash reports by leveraging both standardized fields and unstructured narrative descriptions. CRASH operates on a unified representation of each incident to generate concise summaries, attribute a primary cause, and assess whether the AV materially contributed to the event. Our findings show that (1) CRASH attributes \emph{64\%} of incidents to perception or planning failures, underscoring the importance of reasoning-based analysis for accurate fault attribution; and (2) approximately \emph{50\%} of reported incidents involve rear-end collisions, highlighting a persistent and unresolved challenge in autonomous driving deployment. We further validate CRASH with five domain experts, achieving \emph{86\%} accuracy in attributing AV system failures. Overall, CRASH demonstrates strong potential as a scalable and interpretable tool for automated crash analysis, providing actionable insights to support safety research and the continued development of autonomous driving systems.

\end{abstract}


\section{Introduction}
\label{sec:intro}

The safety premise of Autonomous Vehicles (AVs) remains far off, given the increasing number of failures and fatal incidents. Although simulation and formal safety architectures aim to prevent incidents, rare or catastrophic events can undermine public confidence, erode trust~\cite{blame_trust_simulation}, and provoke debates over accountability and liability~\cite{holland2020verification, pollanen2020blame}.
Reflections on repeated AV missteps warn that without structured learning from past incidents, the industry risks repeating failures~\cite{koopman2024lessons}. Systematic narrative analysis is therefore essential, both for technical improvement and for restoring public trust.
Meanwhile, research on AV crash causality, such as analyses of disengagements and collision classifications, has identified key failure modes, including challenges related to road surface conditions \cite{leilabadi2019depth, favaro2018autonomous}. However, these studies often rely on human-intensive analysis and specialist expertise, which limits their scalability and depth. Emerging tools such as Large Language Models (LLMs) have the potential to transform the analysis of large-scale, unstructured narratives related to safety-critical scenarios \cite{surampudi2024big}, provided their inherent biases are carefully addressed.

This paper introduces CRASH, a reasoning-centric LLM-based agent for scalable analysis of AVs' incident reports. As AV incident databases continue to grow in size and complexity, traditional expert-driven analysis does not scale: manual review is time-consuming, difficult to standardize, and limits the ability to extract cross-incident insights. Rather than replacing human expertise, CRASH reframes the analysis workflow by shifting the primary burden of structured reasoning and synthesis to an LLM-based agent, while retaining humans as reviewers, validators, and decision-makers. In doing so, we demonstrate a new interaction paradigm with large, text-heavy incident report databases that enables systematic, interpretable, and efficient safety analysis.

Our contributions are summarized as follows:
\begin{itemize}
\item A reasoning-driven analysis methodology that operationalizes expert safety reasoning into a structured, multi-step LLM pipeline, enabling consistent interpretation of heterogeneous incident narratives rather than proposing a new data format.
\item A causal attribution agent that performs structured decomposition of incident reports by assigning primary causes, identifying failed AV subsystems, detecting delayed AI perception or response, and generating concise, human-readable summaries to support expert review.
\item A taxonomy-guided analysis framework coupled with automatic aggregation of model outputs, enabling the discovery of recurring AV failure patterns and providing data-driven insights to inform safer system design and policy discussions.
\end{itemize}

By leveraging automated narrative reasoning, we aim to help the AV field become more robust, not only by engineering safer systems but also by institutionalizing learning from failures. The remainder of this paper is organized as follows. Section~\ref{sec:related} reviews related work; Section~\ref{sec:arch} describes our data processing pipeline and details the architecture of our LLM-based agent; Section~\ref{sec:eval} presents evaluation results and it is followed by our findings on the dataset analysis in Section~\ref{sec:findings}; and Section~\ref{sec:conc} concludes with lessons and future directions.

\section{Related Works}
\label{sec:related}
\subsection{Early AV crash analysis} 
\label{early}

Early AV crash analyses (2014--2021) relied on structured, report-level statistical aggregation of California DMV data to characterize disengagement causes, collision types, and attributed fault~\cite{favaro2018autonomous, banerjee2018hands, leilabadi2019depth, alambeigi2020crash, houseal2022causes, CADMV2025}. While effective for trend identification, these approaches operate on predefined categorical labels without systematically inspecting the underlying narrative descriptions for causality chains or system-level interactions, a distinction central to our methodological contribution.

\subsection{Qualitative analysis of reports}

In contrast to purely statistical analyses, works such as \cite{shah2019safe, shokerwip} conducted qualitative examinations of individual crash cases, manually analyzing narrative descriptions to extract contributing factors. By design, these studies rely on close reading and expert interpretation of a limited number of reports, enabling richer contextual insights but restricting scalability and reproducibility. Visualization platforms such as \cite{avcrashes2025} further support exploratory qualitative analysis by curating narratives into searchable and interactive formats. However, these tools remain centered on manual inspection rather than automated, systematic narrative processing. These efforts demonstrate the value of unstructured crash narratives. Nevertheless, they lack a scalable framework for extracting and organizing causal and system-level information across large corpora, an explicit methodological gap addressed in our work.

\subsection{Text classification and summarization using NLP}

Natural language processing (NLP) has increasingly been applied to large-scale transportation safety reports to move beyond purely manual or aggregate analyses. For example, \cite{zhang2022disengagement} proposed an NLP-based pipeline for DMV disengagement reports (2014--2020), automatically labeling disengagement causes using predefined taxonomies and a supervised classification architecture. Their methodology formalizes narrative processing but remains tied to explicit label engineering and fixed output categories. Beyond transportation, \cite{goel2024x} demonstrates that grounding LLM prompts in structured contextual signals improves root-cause recommendations and incident classification, highlighting the importance of domain-specific contextualization for reliable large-scale reasoning. Similarly, \cite{zhang2025improving} benchmarked transformer-based narrative-mining approaches on Kentucky police reports (2015--2022), comparing fine-tuned models such as RoBERTa \cite{liu2019roberta} with zero-shot LLMs including DeepSeek-R1:70B \cite{guo2025deepseek}. Their evaluation emphasizes performance–cost trade-offs and the role of model selection in scalable deployment.

Collectively, these works formalize narrative analysis through supervised learning, fine-tuning, or predefined classification schemes. However, they primarily frame the task as label prediction or severity estimation, rather than structured reasoning over causal chains and multi-stage system interactions. Our work builds on this methodological progression by treating crash narratives as inputs to a reasoning-oriented agent rather than solely a classification task. Using a more recent and comprehensive dataset (2021--2025), we generalize narrative-based analysis at scale without relying on rigid, manually engineered label sets. This enables systematic extraction of system-level failure patterns, bridging the gap between descriptive statistics, case-by-case qualitative analysis, and conventional supervised NLP pipelines. Table~\ref{tab:method_comparison} compares selected prior studies and CRASH across five methodological dimensions explicitly reflected in the table: use of statistical analysis, expert-defined rules, NLP/LLM techniques, scalability, and system-level reasoning. The comparison highlights methodological differences in how incident data are analyzed, and whether approaches move beyond descriptive statistics toward automated, structured reasoning.

\begin{table}[H]
\centering
\small
\caption{Methodological comparison of selected prior work and CRASH.}
\label{tab:method_comparison}
\setlength{\tabcolsep}{5pt}
\begin{tabularx}{\textwidth}{cccccc}
\toprule
\textbf{Work} & \textbf{Stats} & \textbf{Expert Rules} & \textbf{NLP/LLM} & \textbf{Scalable} & \textbf{Sys. Reason} \\
\midrule
Favaro et al. \cite{favaro2018autonomous}   & $\checkmark$ & -- & -- & -- & -- \\
Houseal et al. \cite{houseal2022causes}     & $\checkmark$ & -- & -- & -- & -- \\
Shah et al. \cite{shah2019safe}             & -- & $\checkmark$ & -- & -- & $\checkmark$ \\
Zhang et al. \cite{zhang2025improving}      & $\checkmark$ & -- & $\checkmark$ & $\checkmark$ & -- \\
\midrule
\textbf{CRASH (Ours)}                              & $\checkmark$ & $\checkmark$ & $\checkmark$ & $\checkmark$ & $\checkmark$ \\
\bottomrule
\end{tabularx}
\end{table}
\section{CRASH: Cognitive Reasoning Agent for Safety
Hazards}
\label{sec:arch}

\begin{figure}
    \includegraphics[width=0.7\textwidth]
    {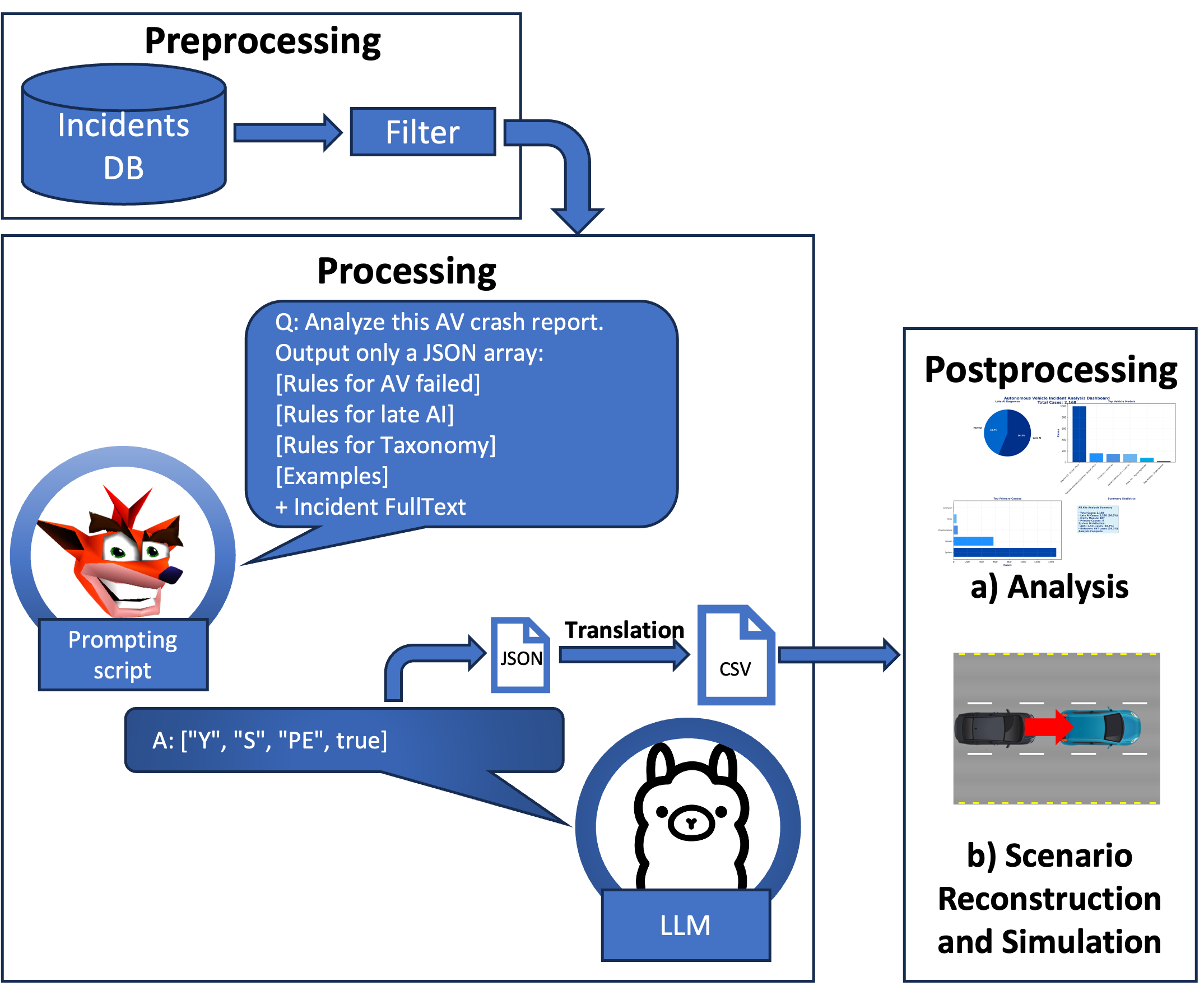}
    \centering
    \caption{CRASH Architecture. During preprocessing, the database is filtered for incomplete entries and unified into four columns. Then, each row is sent to Processing through the LLM. We finally aggregate all information into a CSV file and send it to Postprocessing, where the data is used to generate our analysis and simulation descriptions.}
    \label{fig:architecture}
\end{figure}

To address the gaps identified in Section~\ref{sec:related}, namely the limited scalability of human-driven analysis and the absence of systematic causal reasoning over narrative reports, we introduce the CRASH agent architecture. Rather than treating the LLM as an isolated reasoning component, CRASH is designed as a modular, reproducible processing pipeline that ensures traceability from raw reports to insights into data distribution and simulation-ready outputs. The architecture separates data conditioning, language-based reasoning, and structured interpretation into distinct stages to avoid black-box behavior and preserve analytical rigor. The proposed CRASH architecture, depicted in Fig.~\ref{fig:architecture}, consists of three main modules: \textbf{Preprocessing}, where we digest and filter the incidents database; \textbf{Processing}, where the filtered data are provided to the LLM for structured causal extraction; and \textbf{Postprocessing}, where we analyze the resulting data distributions, LLM outputs for alignment, and generate structured inputs for simulation tools to recreate incident scenarios. The workflow of these modules is described in detail below.

\subsection{Preprocessing}
\subsubsection{Dataset and Filtering}
CRASH is designed to operate on any structured incident report database that provides standardized metadata fields alongside free-text narrative descriptions. For this study, we instantiate the pipeline on the National Highway Traffic Safety Administration (NHTSA) database~\cite{NHTSA2025_SGO_CrashReporting}. This crash reporting program provides a comprehensive, multi-manufacturer, and geographically diverse collection of ADS and ADAS incidents. Alternative sources, such as voluntary fleet summaries~\cite{WaymoImpact2025} or state-level ADS registries~\cite{CADMV2025}, tend to be either self-reported and fleet-specific or limited to a single region, making them less suited for uncovering generalizable system-level trends. The distribution of incidents by reporting entity for the NHTSA dataset is shown in Table~\ref{tab:maker_distribution}.

The original data, distributed across separate CSV files, was merged into a single dataset. Each report was restructured into four columns: \textit{Report ID} (unique identifier), \textit{Reporting Entity/Make} (categorical feature), and a \textit{Full Text} field that concatenates all remaining structured metadata with the original narrative, providing the LLM a unified context window per incident. Entries with redacted or missing narratives were filtered out; pre- and post-filtering distributions are shown in Table~\ref{tab:dataset_stats}.

\begin{table}[ht]
\centering
\caption{Distribution of Incident Cases by Reporting Entity}
\label{tab:maker_distribution}
\begin{tabularx}{\textwidth}{XXcc}
\toprule
\textbf{Model} & \textbf{Make} & \textbf{Cases} & \textbf{\%} \\
\midrule
Jaguar I-Pace & Waymo LLC & 1015 & 46.83\% \\
Cruise Av & GM LLC / Cruise LLC & 355 & 16.37\% \\
Jaguar I-Pace & Transdev & 172 & 7.93\% \\
Honda Civic & Honda & 113 & 5.21\% \\
Toyota Highlander & Zoox, Inc. & 88 & 4.06\% \\
Others & Various & 425 & 19.60\% \\
\midrule
 & \textbf{Total} & \textbf{2168} & \textbf{100.00\%} \\
\bottomrule
\end{tabularx}
\end{table}



\begin{table}[h]
\centering
\caption{Case distribution before and after filtering}
\label{tab:dataset_stats}
\begin{tabularx}{\textwidth}{Xcc}
\toprule
\textbf{Dataset Category} & \textbf{Original Cases} & \textbf{Final Cases}\\
\midrule
ADS & 1790 & 1764 \\
ADAS  & 2582 & 352 \\
Others & 3520 & 52 \\
\midrule
\textbf{Total} & \textbf{7892} & \textbf{2168} \\
\bottomrule
\end{tabularx}
\end{table}


\subsection{Processing}
\subsubsection{LLM prompt construction}
As shown in Fig. \ref{fig:architecture}, our data processing technique comprises a Prompting Script, written in Python, that interfaces with the LLM. The prompt was developed using Prompt Engineering and \textit{In-Context Learning} (ICL) \cite{dong2024survey}, a paradigm in which the model is provided with a specific persona, a constrained rule set, and illustrative examples to guide its output. Preliminary testing showed that this approach outperformed the \textit{Chain-of-Thought} (CoT) method \cite{wei2022chain}. While CoT encourages ``step-by-step'' reasoning, it significantly increased output token length and frequently caused the model to deviate from the required JSON schema due to context-window bottlenecks and ``hallucinated'' conversational filler. The finalized prompt design, detailed in Fig. \ref{final-prompt}, favors constrained classification over open-ended generation to maximize reliability across large datasets. By explicitly defining the ``Rules for AV Failed'' we inject domain-specific expert knowledge directly into the model’s inference path, preventing the LLM from relying solely on its internal, and potentially biased, training weights regarding accident liability. The decision to use one-shot examples was made to anchor the model’s understanding of the short-hand coding system (e.g., \texttt{PE}, \texttt{PL}), which serves as a compression technique to reduce latency and cost. Alternatives such as Fine-Tuning were considered but ultimately rejected due to the ``black-box'' nature of weights and the high computational cost of retraining when new AV failure modes are identified. Similarly, Zero-Shot prompting was found insufficient for the technical nuances of this task, as the model occasionally confused ``stationary'' rear-endings with ``active'' contributions without explicit guidance.

\begin{figure}[H]
\centering
\begin{tcolorbox}[
    enhanced,
    colback=white!5,
    colframe=blue!50!black,
    title=System Prompt: CRASH Agent,
    fonttitle=\bfseries,
    width=0.98\linewidth,
    boxrule=0.5pt
]
\small
\textbf{Role:} You are an autonomous vehicle (AV) incident analyst. Perform all reasoning internally and \emph{only} output the final structured result.

\medskip
\textbf{Tasks}
\begin{itemize}[nosep, leftmargin=1.5em]
    \item 1. Decide if AV contributed. 2. Select primary cause. 3. Identify failed system (if S). 4. Check if AI response was late. 5. Assign secondary cause.
\end{itemize}

\medskip
\textbf{Rules for AV Failed}
\begin{itemize}[nosep, leftmargin=1.5em]
    \item Moving AV action contributed $\rightarrow$ \texttt{Y}
    \item Parked/Stationary rear-ended $\rightarrow$ \texttt{N} (unless avoidable $\rightarrow$ \texttt{Y})
    \item Delayed detection/reaction $\rightarrow$ \texttt{Y} and \texttt{Late AI = true}
    \item Insufficient info $\rightarrow$ \texttt{I}
\end{itemize}

\begin{tabular}{ll}
\textbf{Causes:} & \texttt{S} (Sys), \texttt{H} (Hum), \texttt{E} (Env), N (None) \\
\textbf{Systems:} & \texttt{PE} (Perc), \texttt{PL} (Plan), \texttt{CO} (Control), \texttt{SW}, \texttt{HW}, \texttt{HA}, \texttt{N}\\
\end{tabular}

\medskip
\textbf{Secondary Cause Rules}
\begin{itemize}[nosep, leftmargin=1.5em]
    \item Provide only if multiple factors; Must differ from primary (\texttt{S, H, E, N})
\end{itemize}

\medskip
\textbf{Examples}
\begin{tcolorbox}[colback=gray!10, colframe=gray!30, boxrule=0.5pt, left=2pt, right=2pt, top=2pt, bottom=2pt]
\scriptsize
\textbf{Ex 1:} AV rear-ended while stopped at a red light. \\
\textbf{Output:} \texttt{\{"AV\_Failed": "N", "Cause": "H", "System": "N", "Late": false\}}

\smallskip
\textbf{Ex 2:} AV failed to detect pedestrian; emergency braking engaged 0.5s after impact. \\
\textbf{Output:} \texttt{\{"AV\_Failed": "Y", "Cause": "S", "System": "PE", "Late": true\}}
\end{tcolorbox}
\end{tcolorbox}
\caption{System prompt for the CRASH Agent, incorporating heuristic rules and one-shot examples.}
\label{final-prompt}
\end{figure}

\subsubsection{\textbf{CRASH Taxonomy of AV Incident Causes}}
\begin{figure}
    \centering
    \includegraphics[width=0.8\linewidth]{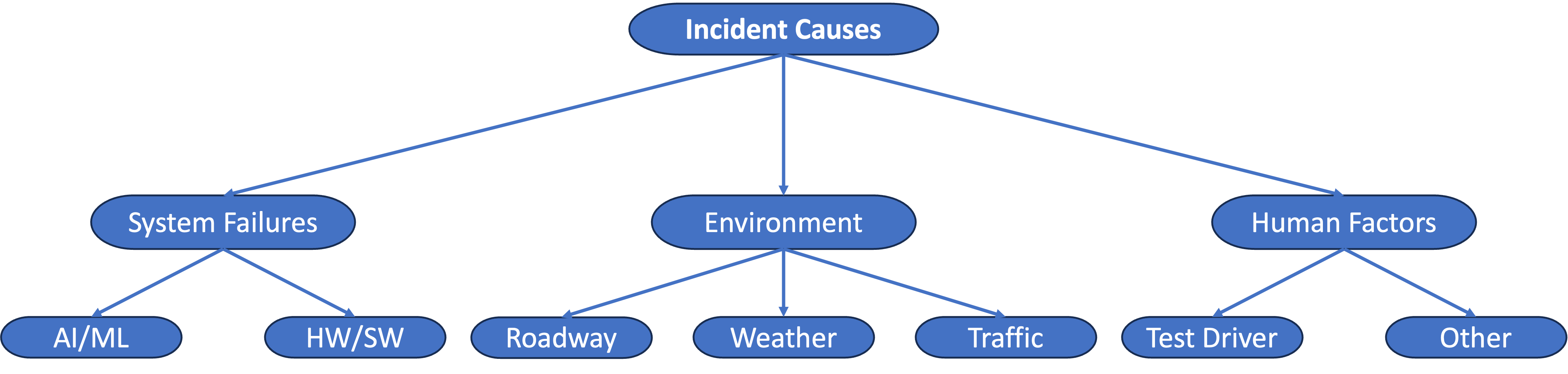}
    \caption{Compact taxonomy of AV incident causes.}
    \label{fig:av-taxonomy}
\end{figure}

To enable consistent causal attribution and downstream simulation reconstruction, we develop a structured taxonomy and standardized output format tailored to AV incident analysis. Existing taxonomies either focus narrowly on disengagement reporting or emphasize high-level behavioral abstractions, limiting their suitability for system-level reasoning and simulation-ready representations~\cite{zhang2022disengagement,saffary2024developing}. In particular, prior categorizations often center on the Sense–Plan–Act (SPA) decomposition without explicitly modeling AI-specific failure modes, cross-module interactions, or environmental and human co-factors. Our taxonomy extends beyond SPA by organizing causes into three unified categories: System Failures, Human Factors, and Environmental Conditions; as depicted in Fig.~\ref{fig:av-taxonomy}. System failures span perception, prediction, planning/control errors, software faults, latency, delayed handover, and hardware and communication issues. Human factors cover both AV operators (e.g., inattention, premature intervention) and other road users (e.g., reckless driving). Environmental conditions include adverse roadway, weather, and complex traffic scenarios. Incidents may involve multiple interacting factors; this study emphasizes primary cause attribution, though secondary causes are also captured.

\subsubsection{Choosing the LLM model}
We evaluated models from the \textit{Qwen3} and \textit{DeepSeek} families~\cite{qwen3github,guo2025deepseek}, deployed locally via \textit{Ollama}~\cite{ollama}, seeking a model whose context window accommodates the full prompt, incident text, and structured response while consistently producing valid JSON. Decoding was fixed at \texttt{temperature = 0} and \texttt{top\_p = 1} for determinism. Models with fewer than 14B parameters proved unreliable for structured output; 7B variants required multi-generation voting, matching the latency of a single 32B run. The best configuration was \textit{DeepSeek-R1} (32B, Q4\_K\_M quantization) deployed locally through Ollama, averaging $\sim$30 seconds per case. Local deployment eliminates API costs and aids reproducibility; the pipeline remains model-agnostic and extensible to cloud-hosted models.

\subsection{Postprocessing and Human-in-the-Loop Validation}
\label{postprocessing}
The postprocessing stage serves a dual role: structuring causal distributions for quantitative analysis and assessing output quality through expert alignment. The standardized taxonomy enables reproducible aggregation and traceable mapping from narratives to structured causal representations. We also incorporate a human-in-the-loop validation protocol: domain researchers evaluated correctness, clarity, and causal consistency of outputs via a structured survey, and their aggregated feedback was used to iteratively refine prompt instructions, output constraints, and attribution guidelines. This lightweight alignment mechanism~\cite{christiano2017deep,ouyang2022training} improves consistency and domain fidelity without updating model weights.

\section{Evaluation}
\label{sec:eval}

We evaluate CRASH as a system-level analysis pipeline to determine whether structured reasoning over unstructured crash narratives can be performed reliably, reproducibly, and at scale. The evaluation is structured across three primary axes:
\begin{enumerate}
    \item \textbf{System reliability:} Validating the consistency and formatting stability of the LLM outputs (Sec.~\ref{sec:eval_perf}).
    \item \textbf{Expert agreement:} Establishing a qualitative ground truth across 50 canonical incidents to measure reasoning accuracy and comparing against traditional NLP baseline heuristics (Sec.~\ref{sec:eval_human}).
    \item \textbf{Runtime efficiency:} Assessing computational inference bottlenecks and scalability compared to manual analysis (Sec.~\ref{sec:eval_runtime}).
\end{enumerate}

\subsection{System Reliability}
\label{sec:eval_perf}

From a systems perspective, CRASH demonstrates stable, structured generation and deterministic behavior under fixed decoding parameters. JSON outputs remain consistent, with formatting failures occurring in only 2\% of cases; these were automatically corrected through a retry mechanism.

\subsection{Expert Agreement}
\label{sec:eval_human}

To assess output quality, five researchers from our group independently reviewed 50 representative cases sampled from the dataset. All evaluators actively work on automotive systems and collectively bring decades of academic and industry experience in autonomous driving and intelligent transportation. Each case received two independent evaluations through a 10-case overlap between reviewers. Evaluators relied exclusively on the \textit{Full Text} field of each report, without external knowledge about manufacturers or prior events. They assessed four dimensions: AV responsibility, late AI response, primary cause, and failed subsystem. For each dimension, they assigned one of three labels: \emph{Correct}, \emph{Incorrect}, or \emph{Insufficient Context}. To evaluate accuracy against these subjective human judgments, we derived a lenient, proxy ``gold label'' dataset. If at least one human evaluator deemed the model's output \emph{Correct}, that output string was accepted as the gold label for scoring purposes. If all evaluating researchers agreed that providing a definitive answer was impossible, it was marked \emph{Insufficient Context}. This criterion reflects realistic investigative settings, where incident narratives naturally contain incomplete or ambiguous information.

Under this derived gold standard, CRASH achieves 86\% accuracy for AV responsibility, 84\% for late AI detection, 76\% for primary cause attribution, and 46\% for failed subsystem identification. The results indicate strong alignment with expert interpretation on high-level causal dimensions. At the same time, subsystem classification remains more challenging because perception, planning, and control failures often interact and cannot be cleanly separated in narrative reports.

\subsubsection{Human Agreement and Dataset Ambiguity}
Reviewer agreement ranged from 53--67\% across dimensions, highlighting the inherent ambiguity of incident narratives. Evaluators frequently assigned \emph{Insufficient Context} labels, particularly for subsystem attribution, where up to 68\% of cases lacked enough information for a definitive judgment. This observation reflects a structural limitation of real-world crash reports: descriptions often omit timing details, sensor behavior, and internal autonomy stack decisions. Consequently, subsystem attribution requires stronger evidence than higher-level causal interpretation, explaining the lower accuracy observed for this dimension.

\subsubsection{Baseline Comparison}
\label{sec:eval_baselines}

To demonstrate whether CRASH genuinely performs contextual reasoning or mimics simpler statistical priors, we evaluate the system against two reference baselines mapped to the traditional NLP methodologies identified in Section~\ref{sec:related}:
\begin{enumerate}
    \item \textbf{Majority Class Data (Descriptive Statistics):} A naïve statistical predictor that blindly outputs the most frequent label observed across the entire 2,168-case dataset (e.g., always predicting \textit{AV Failed: Yes} or \textit{Primary Cause: System}).
    \item \textbf{Keyword Rules (Manual Extraction):} A deterministic system that relies on robust Regular Expression heuristics searching the report's \textit{Full Text}. For instance, finding ``ADAS engaged'' asserts AV failure, while counting terms like ``rain'' or ``weather'' asserts an environmental cause.
\end{enumerate}

To ensure fairness, both baselines were scored against the same proxy gold labels generated from the human evaluation. If reviewers deemed a report to have \emph{Insufficient Context}, baseline answers automatically scored zero, penalizing them equally alongside CRASH for guessing unanswerable queries. Similarly, if human reviewers marked CRASH's output as incorrect but lacked sufficient detail to formulate an opposing ground truth label, baseline predictions were conservatively penalized. This architecture mirrors the absolute constraints under which CRASH was evaluated without artificially inflating baseline difficulty. Table~\ref{tab:baselines} contextualizes the resulting performance.

\newcolumntype{C}{>{\centering\arraybackslash}X}
\begin{table}[t]
\caption{Baseline comparison on the 50 expert-evaluated cases (lenient criterion).}
\label{tab:baselines}
\centering
\begin{tabularx}{\textwidth}{X|CCCC}
\toprule
\textbf{Method} & \textbf{AV Fail} & \textbf{Late AI} & \textbf{Cause} & \textbf{Sys. Fail} \\
\midrule
Majority class        & 54\% & 34\% & 42\% & 22\% \\
Keyword rules         & 48\% & 46\% & 44\% & 18\% \\
\rowcolor{gray!10}
\textbf{CRASH} (ours) & \textbf{86\%} & \textbf{84\%} & \textbf{76\%} & \textbf{46\%} \\
\bottomrule
\end{tabularx}
\end{table}

CRASH consistently outperforms both baselines across all dimensions. 
Relative to the strongest baseline for each task, the system reduces 
classification error by approximately 70\% for AV responsibility and 
late AI detection, 57\% for primary cause attribution, and 31\% for 
subsystem identification. Because the majority predictor captures dataset priors and the keyword heuristic reflects simple lexical pattern matching, the improvement over 
both baselines suggests that CRASH performs structured reasoning over incident narratives rather than relying solely on surface cues. Taken together, these results show that CRASH achieves strong alignment with expert interpretation on high-level causal dimensions while remaining appropriately conservative when reports lack sufficient technical detail. This behavior suggests that the system captures meaningful causal structure in crash narratives rather than relying solely on lexical patterns.


\subsection{Runtime Efficiency}
\label{sec:eval_runtime}

Inference constitutes the primary computational bottleneck, with an average processing time of approximately 30 seconds per report on two NVIDIA A4500 GPUs (40GB VRAM). Even with this inference cost, the system processes incident reports substantially faster than manual expert review, which often requires several minutes and repeated readings of the narrative.

\section{Findings on AD Safety Incidents in NHTSA Data}
\label{sec:findings}

\begin{figure*}[t]
\centering
\includegraphics[width=0.65\textwidth]{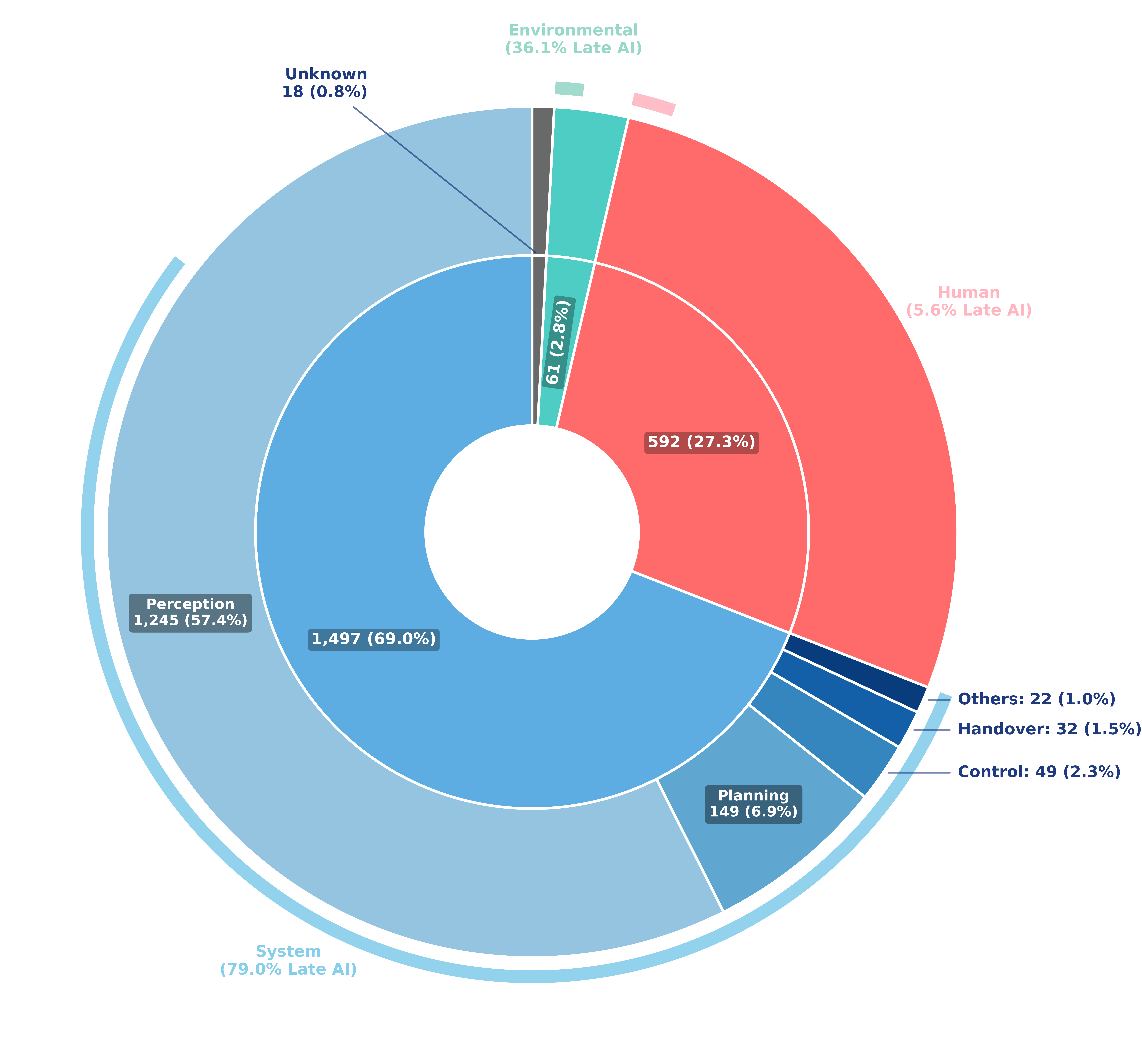}
\captionsetup{font=small}
\caption{Primary cause distribution and subsystem breakdown across 2,168 AV incidents.}
\label{fig:all-primary-causes}
\end{figure*}

We summarize in Fig. \ref{fig:all-primary-causes} the principal quantitative trends extracted from 2,168 NHTSA incident reports using CRASH. System-related failures, shown in tones of blue, dominate the dataset, accounting for 1,497 incidents (69\%), far exceeding human factors (red, 27.3\%) and environmental causes (green, 2.8\%). This dominance is consistent with prior studies showing that perception errors and scene interpretation remain major challenges in complex urban environments. Within system-related incidents, perception failures dominate (1,245 cases), followed by planning (149), control (49), and handover (32). This trend is expected because perception modules must interpret highly variable real-world scenes, including occlusions, unusual vehicle behavior, and complex urban environments, which remain challenging even for modern sensor fusion pipelines. Hardware and software malfunctions occur only rarely, suggesting that most failures stem from algorithmic limitations in scene interpretation rather than physical component faults.

Late AI responses, shown by the outer ring in the image, appear in 57.1\% of all reports and are strongly coupled with system failures: 79\% of system-level failure cases occur alongside late responses. This observation is consistent with the tight real-time constraints of AV stacks, where perception delays propagate to planning and control modules, reducing the available reaction time. This pattern suggests that latency frequently amplifies localized perception or decision errors into full collision outcomes, highlighting timing as a structural constraint in AV safety.

\begin{wrapfigure}{r}{0.45\textwidth}
\centering
\includegraphics[width=\linewidth]{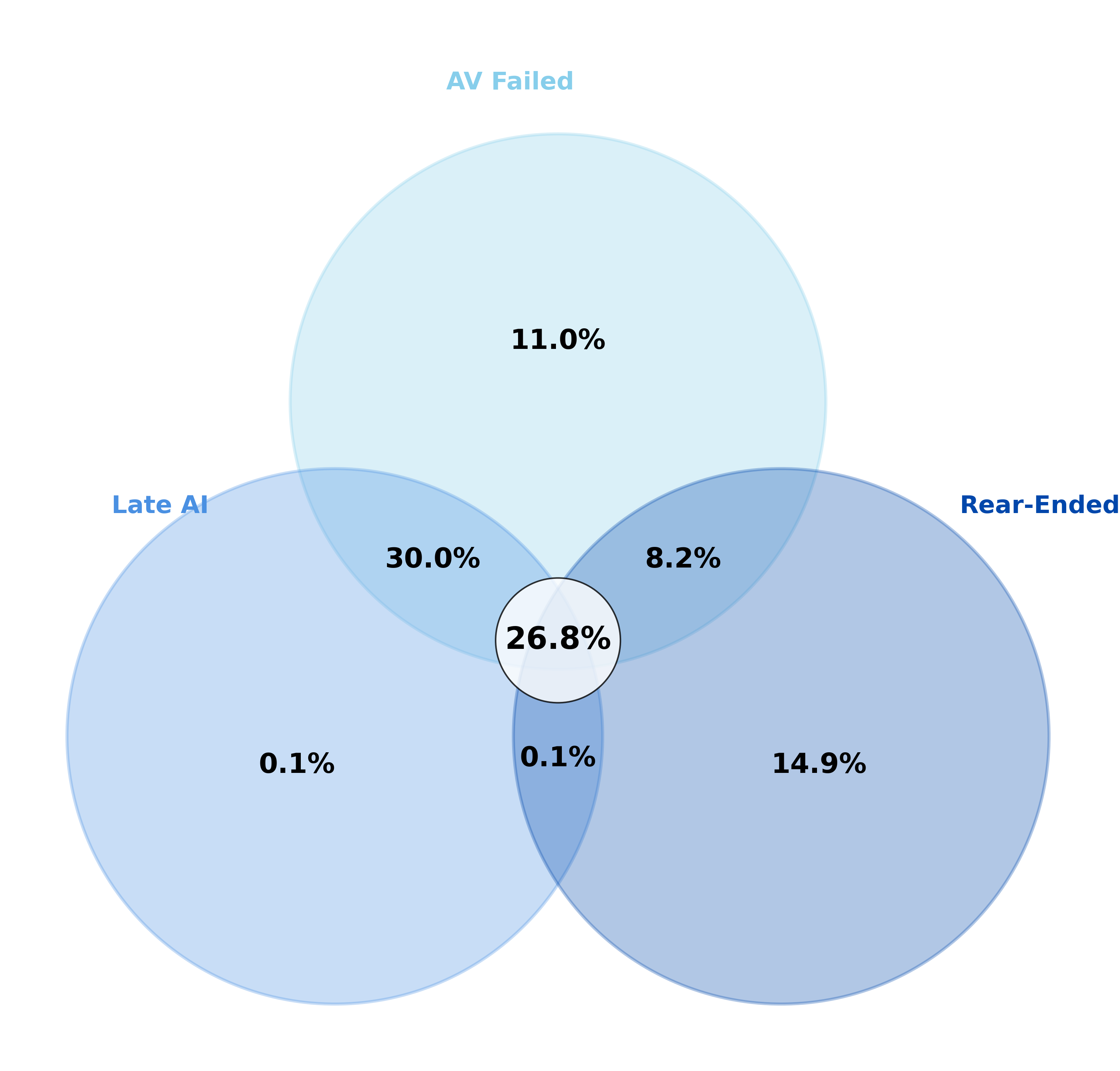}
\captionsetup{font=small}
\caption{Overlap between late AI behavior, AV failures, and rear-end collisions ($N{=}2{,}168$).}
\label{fig:venn}
\end{wrapfigure}

To further examine timing-related failures, we analyze how delayed AI behavior intersects with AV failures and collision types. Rear-end collisions account for approximately half of all incidents in the dataset. Rear-end collisions are common in mixed-autonomy environments because AVs often adopt conservative braking policies that may not align with the expectations of human drivers following behind. While prior studies often report that surrounding drivers predominantly rear-end AVs, our structured analysis identifies 583 (26.8\%) rear-end cases in which delayed perception or decision-making plausibly contributed to the outcome. As illustrated in Fig.~\ref{fig:venn}, delayed detection or reaction substantially overlaps with both AV failures and rear-end collisions, reinforcing the interpretation of latency as an amplifying factor rather than an isolated fault category.

Overall, these findings indicate that perception limitations and timing-related degradation structurally dominate AV safety incidents. The dataset-level trends complement the system-level evaluation by showing how structured reasoning at scale reveals recurring latency-driven constraints within the autonomy stack.

\section{Conclusion}
\label{sec:conc}

This work presented CRASH, a reasoning-centered agent designed to structure and interpret AV incident reports at scale. By decomposing crash narratives into meaningful causal dimensions and aligning its outputs with expert judgment, CRASH enables transparent attribution of responsibility and consistent system-level analysis. Both qualitative and quantitative evaluations show that the agent reliably captures high-level causes while remaining cautious in ambiguous cases. The human review process also highlighted the cognitive effort required to manually analyze incident reports, with evaluators frequently rereading cases to reach consistent conclusions. In contrast, CRASH processes each report in roughly 30 seconds on modest hardware, producing coherent and context-aware reasoning in a fraction of the time. 

Beyond efficiency gains, CRASH reveals important safety patterns. Most notably, the prevalence of timing-related failures, such as rear-end collisions, suggests that perception and planning latency may constitute a critical, underexamined bottleneck in AV safety. By systematically identifying such cross-cutting trends, CRASH moves incident analysis beyond descriptive summaries toward actionable, system-level insights. As with any LLM-based system, semantic hallucination remains a risk: the model may plausibly attribute causes not fully grounded in the narrative. The constrained output schema, domain-specific rules, and deterministic decoding mitigate this by limiting generative freedom, but systematic validation on larger annotated subsets is a priority for future work. Overall, this work demonstrates that LLM-based reasoning agents can meaningfully augment safety auditing in complex, text-intensive domains. By combining interpretability, adaptability, and scalable analysis, CRASH provides a practical foundation for future research on automotive perception, safety validation, and human--AI collaboration in safety-critical systems.

%
\bibliographystyle{splncs04}
\bibliography{ref}
\end{document}